# Quasi-Bayesian Strategies for Efficient Plan Generation: Application to the Planning to Observe Problem

Fabio Cozman            Eric Krotkov
Robotics Institute, School of Computer Science — Carnegie Mellon University
Pittsburgh, PA 15213

## Abstract

Quasi-Bayesian theory uses convex sets of probability distributions and expected loss to represent preferences about plans. The theory focuses on decision robustness, i.e., the extent to which plans are affected by deviations in subjective assessments of probability. Generating a plan means enumerating the actions to be taken and providing information about the robustness of the actions. The present work presents plan generation problems that can be solved faster in the Quasi-Bayesian framework than within usual Bayesian theory. We investigate this on the planning to observe problem, i.e., an agent must decide whether to take new observations or not. The fundamental question is: How, and how much, to search for a "best" plan, based on the precision of probability assessments? Plan generation algorithms are derived in the context of material classification with an acoustic robotic probe. A package that constructs Quasi-Bayesian plans is available through anonymous ftp.

## 1 INTRODUCTION

Agents choose a plan of action by comparing its possible outcomes against the outcomes of other plans. Bayesian theory suggests that the basis for such comparisons is expected loss with a single probability distribution. Quasi-Bayesian theory, as axiomatized by Giron and Rios [Giron and Rios, 1980], also relies on expected loss, but uses a convex set of probability distributions to represent the agent's beliefs. Many scholars agree that assuming an agent uses a single probability distribution is too restrictive [Breese and Fertig, 1991; Levi, 1980; Shafer, 1987]. But there has been little agreement on how to make decisions with several distributions; many seem to think that theories with several distributions will always lead to intractable decision making problems.

Recently, great attention has been given to Robust Bayesian Statistics, which uses Quasi-Bayesian sets of distributions to represent imprecision of subjective probability assessments [Berger, 1985; Walley, 1991]. A *robust* decision is one that can be safely taken despite the imprecision in the probability assessments; a *non-robust* decision is one that may produce wildly different results depending on the adopted distribution.

In this paper, we explore a Quasi-Bayesian approach to plan generation. Generating a plan means enumerating the actions to be taken and providing information about the robustness of the actions. Our approach puts less emphasis on the search for unique "best" decisions than the usual Bayesian approach. Essentially, the agent is required to choose admissible decisions and to monitor and report robustness of these decisions. We clarify the terms involved in this requirement in sections 2 and 3.

The central point of this work can be expressed by a short fable. Suppose two archers try to hit a target (Figure 1). The first archer, a Bayesian, considers that hitting the center of the target is the only satisfactory result and orders a new, expensive bow. But the judge only detects if an archer hits the hatched region. If both archers hit the hatched region, the judge considers them tied and calls other procedures to solve the dispute. That does not prevent the Bayesian archer from trying to hit the center. The second archer, a New Quasi-Bayesian archer, tries simply to reach the hatched circle with a cheap bow. The New Quasi-Bayesian strategy seems wiser *given the lack of precision of the target.*

Our main contribution is to show that the Archers Fable can be formalized for the planning to observe problem. The analogy here is that a point in the target corresponds to a distribution: the Bayesian agent has one, the Quasi-Bayesian agent has many. The Bayesian seeks an answer to the question, how to create an optimal sequence of actions? Such a question is very demading computationally. The Quasi-Bayesian is attentive to the limitations of real probability assessments and seeks an answer to the question, how to create a sequence of admissible actions and quantify the robustness of such actions? The surprising result is that we can answer the latter question without examining the full solution for the former question. To illustrate the details of our solution, we apply it to the planning to observe problem for material classification with a robotic probe.

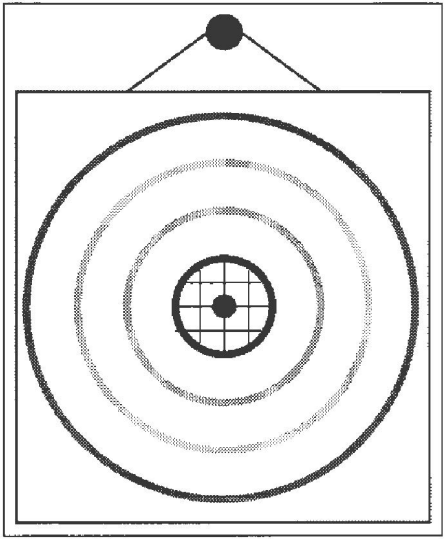

Figure 1: The Archers Fable

## 2   THE QUASI-BAYESIAN FRAMEWORK

Consider this problem: an agent must choose a plan $a_i$ before the state of the world is known; after the state is revealed to be $\theta_j$, the agent pays a loss $l_{ij}$. The losses indicate the preferences of the agent. How should the agent compare two plans, $a_1$ and $a_2$? Quasi-Bayesian theory asserts that there is a nonempty convex set $K$ of probability distributions which summarizes the agent's beliefs. The set $K$ is such that, for plans $a_1$ and $a_2$, $a_1$ is at least as preferred as $a_2$ iff $E[l_{1j}] \geq E[l_{2j}]$ for every probability distribution $K$, where $E[\cdot]$ denotes expected loss. Giron and Rios present a set of axioms that validate the preferences of a Quasi-Bayesian agent [Giron and Rios, 1980]. The set of probability distributions $K$ is called the *credal* set [Levi, 1980]. The representation of preferences conditional on a state is characterized by a convex set of posterior distributions obtained through application of Bayes rule to each one of the distributions in the set of priors[1].

There are other methods for creating sets of probability distributions: inner and outer measures [Good, 1983; Halpern and Fagin, 1992; Ruspini, 1987; Suppes, 1974], intervals of probability (commonly generated by lower probability) [Breese and Fertig, 1991; Chrisman, 1995; Fine, 1988; H. E. Kyburg Jr., 1987;

[1] An introduction to technical aspects of Quasi-Bayesian theory, with a larger list of references, can be found at http://www.cs.cmu.edu/~fgcozman/qBayes.html.

Halpern and Fagin, 1992; Smith, 1961], lower expectations [Walley, 1991]. The belief functions used in Dempster-Shafer theory [Ruspini, 1987; Shafer, 1987] have different interpretations but can be represented as sets of probabilities. Quasi-Bayesian models generalize these ideas. Given a Quasi-Bayesian convex set of probability distributions, a probability interval can be created for every event $A$ by defining lower and upper bounds:

$$\underline{P}(A) = \inf_{P \in K} P(A) \qquad \overline{P}(A) = \sup_{P \in K} P(A).$$

In a different direction, more general models than the Quasi-Bayesian one can be created, for instance theories of decision which use simultaneous sets of losses and probabilities [Levi, 1980; Seidenfeld, 1993].

There are some basic reasons for adopting a Quasi-Bayesian model [Seidenfeld and Wasserman, 1993]. First, Quasi-Bayesian theory builds a realistic account of the imperfections in an agent's beliefs. It can be used to represent poor elicitation of preferences and situations of indifference among actions. Second, robustness studies can be formalized through this model [Berger, 1985]. Third, the theory can represent the disparate opinions of a group of agents [Levi, 1980].

## 3   BUILDING A NEW APPROACH TO DECISION-MAKING

A Bayesian agent can always say that a plan is better than, worse than, or equal to another plan. A Quasi-Bayesian agent may be in a different situation. Consider two plans, $a_1$ and $a_2$, and two distributions $p_1$ and $p_2$ in the credal set. Suppose plan $a_1$ has smaller expected loss than plan $a_2$ with respect to a probability distribution $p_1$, but $a_2$ has smaller expected loss with respect to another probability distribution $p_2$. In this case, $a_1$ and $a_2$ are not comparable by expected loss; both are admissible. What should be done?

Reactions to this question vary. Fertig and Breese, in their work with interval probabilities, simply report all admissible plans [Breese and Fertig, 1991; Fertig and Breese, 1990]. This leaves the actual actions unspecified. Levi argues that plans should not only be admissible, but also be optimal with respect to some distribution in the credal set. He calls such a plan E-admissible [Levi, 1980]. Since there may be several E-admissible plans, Levi suggests secondary guidelines that enforce "security". Others have suggested the agent should minimize the maximum possible value of expected loss, an approach common in Robust Bayesian Statistics under the name of $\Gamma$-minimax [Berger, 1985].

We suggest that Quasi-Bayesian strategies should *specify the admissible decisions and allow the agent to monitor the robustness of such decisions*. These are the two requirements on a plan. There should be no artificially enforced preference among admissible plans:

any admissible plan provide useful guidance if an action must be chosen. Robustness should always be monitored; what use is a "best" plan if it is based on a skewed set of assumptions? As long as a plan provides a method for the detection of non-robust situations, the agent can pick the first admissible decision that admits convenient manipulation in the time available for decision-making. We call this the New Quasi-Bayesian strategy.

The strategy above contains important elements of decision-making as it must be exercised by finite, bounded agents. The agent is required to produce an admissible answer as quickly as possible, and have that as a default solution, as usually required in anytime planning. Further, the agent is required to detect the situations that require additional computation and refinement: those are the non-robust situations.

Compared to the Bayesian strategy, the New Quasi-Bayesian strategy has some remarkable differences. The Bayesian strategy will always be appropriate if there is total confidence on the precision of probability assessments. If that is not the case, the Bayesian strategy calls for a decision analysis of the value of further computation and/or introspection [Heckerman and Jimison, 1989; Horvitz, 1989; Matheson, 1968; Russell and Wefald, 1991]. Such meta-analysis requires probabilities over probabilities, which may be harder to elicit than a simple set of bounds on distributions.

So far we have specified the New Quasi-Bayesian strategy, but it is still unclear how we can use this strategy in any decision problem. In order to do so, we must be able to quickly generate actions and monitor robustness. Ideally, we must be able to do so faster than the usual Bayesian solution, which involves generating actions and either checking the sensitivity of such actions or checking the meta-analysis for those actions. In the remainder of the paper, we show that these goals are met for the **planning to observe** problem with Gaussian measurements. This is a classic Markov decision problem; although we describe the solution for univariate data, the ideas readily extend to multivariate data.

## 4   PLANNING TO OBSERVE WITH GAUSSIAN MEASUREMENTS

We now demonstrate our approach to decision-making on the **planning to observe** problem described as follows:

A series of independent real-valued observations $X_i$ is available to an agent; each observation costs $c$ units of loss and is normally distributed with known variance $1/r$ and unknown mean $\theta$. We indicate this by $X_i \sim N(x_i; \theta, 1/r)$. The agent wants to know whether $\theta$ is larger or equal, or smaller than zero. At any point, the agent can take a new observation or stop and decide: **Smaller** ($d_0$) and **Larger** ($d_1$). When a decision is made, the loss $L(\theta, d_i)$ is defined by Table 1.

Table 1: Losses

|       | $\theta \leq 0$ | $\theta > 0$ |
|-------|-----------------|--------------|
| $d_0$ | 0               | $\theta$     |
| $d_1$ | $-\theta$       | 0            |

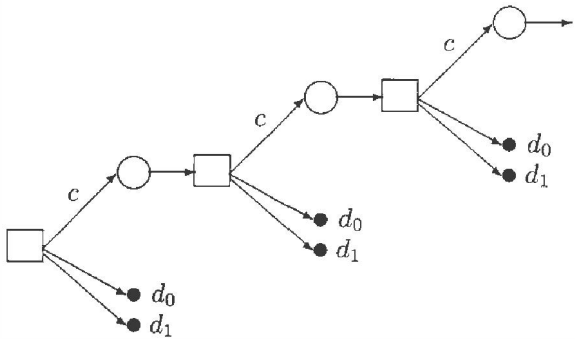

Figure 2: Planning to Observe with Gaussian Measurements

Figure 2 illustrates the dynamics of the problem. At any point, the agent is facing the same question as to whether a decision should be made or an observation should be taken. We want to find a sequence of actions for the agent.

### 4.1   THE CREDAL SET

Prior beliefs about $\theta$ translate into a convex set of probability distributions, the credal set. A realistic model for prior beliefs would take a credal set large enough to represent the non-specificity of the agent's beliefs. Consider the convex set of distributions composed of Gaussian distributions with mean between $\mu_1$ and $\mu_2$, and variance $1/\tau$:

$$\mathcal{N}_0 = \{\text{Convex Combinations of } N(\theta; \mu_j, 1/\tau),$$
$$\mu_j \in [\mu_1, \mu_2]\}. \quad (1)$$

To create a convex set of distributions, it may be necessary to use convex combinations of distributions[2]. To build some intuition, consider the semi-plane $(\tau, \mu)$, $\mu > 0$ (inverse of variance × mean). A Gaussian distribution can be mapped to a point in this semi-plane. The Gaussian distributions in the credal set above can be mapped to a vertical segment of line at $\tau$. Call $\Delta = |\mu_2 - \mu_1|$ the width of the credal set. After each measurement we use Bayes rule to update each distribution in the credal set; after $n$ measurements $x_i$ (with mean $\bar{x}$), the posterior credal set is [Giron and Rios, 1980]:

$$\mathcal{N}_n = \{\text{Convex Combinations of }$$
$$N\left(\theta; \mu_j, (\tau + nr)^{-1}\right),$$

---
[2] A convex combination of a set of functions $\{f_j\}_{j=1}^J$ is given by $\sum_{j=1}^J \alpha_j f_j$, where $\alpha_j$ are non-negative numbers that sum to unity.

$$\mu_j \in \left[\frac{\tau\mu_1 + nr\overline{x}}{\tau + nr}, \frac{\tau\mu_2 + nr\overline{x}}{\tau + nr}\right]\right\}. \quad (2)$$

A fundamental property is that the width of the set shrinks to $\Delta\tau/(\tau + nr)$ after $n$ measurements.

### 4.2 E-ADMISSIBLE PLANS

The New Quasi-Bayesian objective is to find an E-admissible plan and monitor robustness. E-admissible plans for "pure" Gaussians in the credal set will be convenient since the prior and the likelihood are then conjugate [Berger, 1985]. For each Gaussian distribution in the credal set $\mathcal{N}_0$, an E-admissible plan can be generated as follows:

Take the semi-plane $(\tau, \mu)$, $\mu > 0$ (inverse of variance × mean). Divide the plane into three decision regions: a **Continue** region, a **Stop0** region and a **Stop1** region. The posterior density after measurement $x_n$ is $\mathcal{N}_n$, represented by a point in the semi-plane $(\tau, \mu)$. The plan is: check whether the posterior $\mathcal{N}_n$ is in **Continue**, **Stop0** or **Stop1**, and respectively take a new measurement, stop and pick $d_0$, stop and pick $d_1$. The plan is determined by the decision regions, which are created by dynamic programming (value iteration algorithm) [DeGroot, 1970], as explained in Appendix A.

### 4.3 PLANNING WHILE ACTING

The agent has a prior credal set defined by $\mu_1$, $\mu_2$ and $\underline{\tau}$; as measurements are collected, the decision regions must be constructed for $\underline{\tau} + nr$, where $n$ starts from zero.

The whole plan is defined by the upper and lower boundaries of the **Continue** region. The value iteration algorithm essentially brackets this region and converges to the correct boundary, but every iteration requires more effort than the previous iteration. Given any finite amount of computation, the agent, Bayesian or Quasi-Bayesian, has a chart similar to Figure 3 [DeGroot, 1970]. There is an **Indeterminate** region, yet to be explored. The agent can shrink the size of the **Indeterminate** region at high computational cost, as discussed in Appendix A. A real agent has a region of *computational indeterminacy*: because a finite amount of effort is available, not all plans can be evaluated.

We have returned to the Archers Fable. We have the plane $(\tau, \mu)$, and we must hit the boundary of the **Continue** region.

The Bayesian archer wants to find a single curve. Any lack of precision in the prior models will require a sensitivity analysis or a meta-analysis. This may lead the Bayesian archer to spend a vast amount of computation if the archer is considering a point close to the boundary between **Continue** and **Stop** regions.

The New Quasi-Bayesian archer thinks differently. The New Quasi-Bayesian agent recognizes that as soon as there is a point in some region outside the **Indeterminate** region, the whole situation is characterized. If the credal set is inside the **Stop0** region, then stop and pick $d_0$; if the credal set is inside the **Stop1** region, then stop and pick $d_1$; if the credal set is inside the **Continue** region, then take another observation. But if the credal set intersects more than one region, *a non-robust situation has been detected*. The agent has an E-admissible action that can be chosen, but robustness has failed. So the New Quasi-Bayesian archer shrinks the **Indeterminate** region only if *all* distributions in the credal set are inside this region.

To make this strategy concrete, consider an example. Figure 3 shows the four decision regions. Each vertical segment represents a set of "pure" Gaussian distributions with the same variance. The first band is the prior band. Since the band is inside the **Continue** region, the agent takes a new measurement without further computation. Now the band crosses both **Continue** and **Stop1**. The agent knows, *without further computation*, that both taking a new measurement and choosing $d_1$ are admissible, no matter how much additional effort is spent. If an anytime decision is needed at this point, a new observation is taken. But the correct analysis is that the prior imprecision has created a non-robust situation where a possible action is to continue observing.

This strategy links the computational indeterminacy of the planning algorithm to the credal indeterminacy of the agent, formalizing a connection between search effort and model building. There are limits to the effort that is worthy spending in search for a given level of imprecision in a probability assessment. This work appears to be the first analysis of this trade-off with the tools of Quasi-Bayesian theory. Some new questions emerge. First, what are the methods that define the agent's behavior when two decisions are computable and admissible? Second, what are the approximation algorithms (value iteration in this case) that admit a relationship between computational and credal indeterminacies? Third, is the approximation algorithm biased, i.e., is it causing the agent to pick some regions more than others? This happens, for example, if the **Indeterminate** region greatly extends into one of the **Stop** regions but not into the other[3].

### 4.4 PLANNING IN ADVANCE

An analysis of this problem must take into account the possibility that the agent pre-computes the decision regions. In general, suppose the agent wants to pre-compute the relevant decision regions for prior width $\underline{\Delta}$ and prior inverse variance $\underline{\tau}$. So the agent must pre-compute the regions for all $\underline{\tau} + nr$, where $n$ starts from zero.

The New Quasi-Bayesian archer has simply to guar-antee that the band of "pure" Gaussians does not fall entirely inside the **Indeterminate** region. This can be done in a finite number of iterations given the monotone character of the value iteration algorithm (Appendix A) and the fact that the posterior set of Gaussians has a width that decreases as $\Delta\tau/(\tau + nr)$.

---
[3]The fact that this may occur was suggested to us by Prof. T. Seidenfeld.

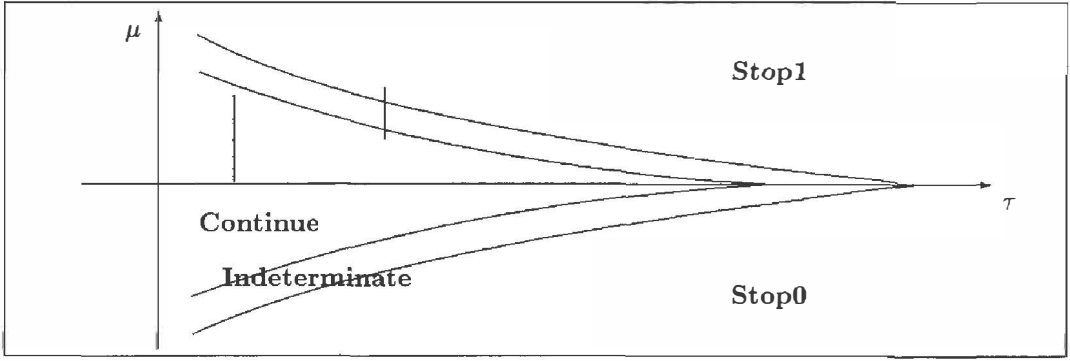

Figure 3: Planning to Observe with Gaussian Measurements and Finite Computational Resources

For the New Quasi-Bayesian archer, the question arises as to whether or not we can determine the boundaries of the **Indeterminate** region without even iterating the value iteration algorithm. Of course, this must depend on the characteristics of the initial credal set. Suppose the agent starts with a given prior variance $1/\underline{\tau}$ and a given prior width $\underline{\Delta}$. The following new result characterizes the situations that admit direct solutions for the boundaries[4]. The result is even stronger than what we just required: it covers *all* plans where priors have width smaller than $\underline{\Delta}$ and variance smaller than $1/\underline{\tau}$.

**Theorem 1** *If $z\Omega(z\underline{\Delta\tau}/2) < c$ for $\tau''^{-1/2} \leq z \leq \tau'^{-1/2}$, then the following decision regions completely specify the agent's behavior for $\underline{\tau}$ and $\underline{\Delta}$:*

**Continue** $|\mu| \leq \dfrac{\mho\left(c\sqrt{\frac{r(\tau+r)}{r}}\right)}{\sqrt{\frac{r(\tau+r)}{r}}}$

**Stop1** $\mu \geq \max(0, \mho(c\sqrt{\tau})/\sqrt{\tau})$

**Stop0** $\mu \leq \min(0, -\mho(c\sqrt{\tau})/\sqrt{\tau})$

**Indeterminate** *otherwise.* □

We can summarize the New Quasi-Bayesian strategy for generating plans:

1. Theorem 1 verifies whether the Quasi-Bayesian plan can be stored in closed-form.

---
[4]The following definitions are used: $\phi(s)$ is the standard Gaussian density and $\Phi(s)$ is the standard Gaussian distribution function; $\Omega(s) = \phi(s) + (1 - \Phi(s))s$; $\mho(s)$ is the inverse function of $\Omega(s)$; $\tau' = \sqrt{(r/2)^2 + (r/2)/(\pi c^2)} - (r/2)$ and $\tau'' = 1/(2\pi c^2)$.

2. If not, then the value iteration algorithm shrinks the **Indeterminate** region. When the **Indeterminate** region is smaller than $\underline{\Delta\tau}/\tau$ for every $\tau$ larger than $\underline{\tau}$, the decision regions are defined.

We now look at a situation that occurs in practice and leads to increased savings within the New Quasi-Bayesian framework. Suppose the agent has to provide plans for a variety of values of the cost of observations $c$. A situation where this happens is illustrated in the next section. Here we consider the costs $c_i$ to belong to a finite set of values $\{c_1, c_2, \ldots, c_m\}$.

The boundaries of the **Indeterminate** region must be generated for each one of the costs $c_i$. The following result is useful:

**Theorem 2** *If the conditions of Theorem 1 are satisfied for a value $c^*$, they are satisfied for a value $c$ larger than $c^*$.* □

At first, the New Quasi-Bayesian identifies a value of $c_i$ that admits closed-form plans using Theorem 1; for larger values of $c_i$ the plans can be directly stored. For other values of $c_i$, the agent must construct boundaries for the decision regions by iteration. The value iteration algorithm must shrink the **Indeterminate** region until it is smaller than the width of the credal set. Again, as the problem became more involved, the savings in the New Quasi-Bayesian scheme increased when compared to the Bayesian prescription.

### 4.5 EVALUATING THE SOLUTION IN A REAL PROBLEM

Consider the construction of a robotic probe for classification of material based on acoustic signals [Krotkov and Klatzky, 1995]. The taks is for a robot to decide whether a material belongs to one of two classes based on the tangent of the angle of internal friction, $\tan\phi$, which is captured from acoustic analysis of impact sounds. This is equivalent to deciding whether a variable $\theta$ (linearly related to $\tan\phi$) is larger or smaller than zero. The losses are given by table 1. The robot is used for a variety of tasks; when the robot is assigned to a task, a cost for robot operation is assigned based on the number of waiting tasks in a queue. So the act of striking a material costs a quantity $c_i$ which belongs to a finite set of possible costs $\{c_1, c_2, \ldots, c_m\}$, corresponding to the size of the queue. Once the task is initiated with a cost $c_i$, the cost remains fixed during that task.

Suppose we want to distinguish metals with $\tan \phi$ above -11 (aluminum has $\tan \phi$ of approximately -2) from non-metallic materials with $\tan \phi$ below -11 (plastic has $\tan \phi$ of approximately -20). We translate these values so that $\theta$ is zero when $\tan \phi$ is -11; now we decide whether $\theta$ is larger or smaller than zero. Experiments suggest a Gaussian model for the measurements: $X_i \sim N(x_i; \theta, r)$, with $r = 1$ [Krotkov and Klatzky, 1995].

Very sparse knowledge about $\theta$ must be translated into a belief model. Trying to model this with a single prior leads to a number of arbitrary choices. Instead, take the prior model to be a Quasi-Bayesian set $\mathcal{N}_0$ with variance $1/\underline{\tau} = 4$, $\mu_1 = -5$, $\mu_2 = 5$ (so that $\underline{\Delta} = 10$).

The last element to be specified is the cost of an observation. We consider a vector of possible costs, depending on the state of the robot, and assume a linear relation: $c_i = ci$, $i \in \{1, \ldots, 10\}$, where $i$ is the number of tasks in the robot queue, including the one the robot is operating on. We must define $c$, the cost of an observation when no task is waiting. Instead of fixing a value of $c$ arbitrarily, we build some intuition by asking the question: If we had just a single Gaussian prior defined by mean $\mu$, $\mu > 0$, and variance $1/\underline{\tau}$, and the right to take a single observation, what would we do? For $\mu$ larger than a certain value $\mu'$, we would rather take $d_1$ than pay $c$. So our choice of $\mu'$ encodes the value of $c$. The value $c$ such that $\mu'$ is the boundary of the **Continue** region for $\underline{\tau}$ is:

$$c = \frac{\Omega\left(\sqrt{\frac{\tau(\tau+r)}{r}}\mu'\right)}{\sqrt{\frac{\tau(\tau+r)}{r}}}.$$

In our particular example, we took $\mu' = 4$, signifying that, unless we believed strongly that $\theta > 0$, we would prefer to take a new observation. In other words, we regard the cost $c$ to be relatively small. The use of the previous equation with $\underline{\tau} = 0.25$, $\mu' = 8$ and $r = 1$ gives us $c = 7.88 \times 10^{-3}$. We round that and adopt $c = 0.01$ to represent an appropriate cost for the measurements, so we have $c_i \in \{0.01, 0.02, \ldots, 0.1\}$.

First we search for closed-form solutions. Theorems 1 and 2 indicate that values of $c > 0.081$ lead to closed-form plans. We must only obtain plans for $c_i \in \{0.01, 0.02, \ldots, 0.08\}$ (for the results in this section, we used a symbolic package which we are making publicly available; see Appendix B for details).

For $c_i = 0.07$, the **Indeterminate** region is shrunk sufficiently by a single iteration of the value iteration algorithm (i.e., reaching $\tilde{\rho}'_2$ and $\tilde{\rho}''_2$). For $c_i = 0.03$, two iterations of the algorithm are necessary to obtain the decision regions. Below this value, the computational effort involved in computing the decision regions is rather large. Instead of sacrificing time to compute these plans, we use "lumped" observations. When $c_i \in \{0.01, 0.02\}$, we take two observations at a time; the "lumped" observation is the average of the two observations and has precision 0.5. The bottom of Figure 4 shows the resulting decision regions for $c_i = 0.02$.

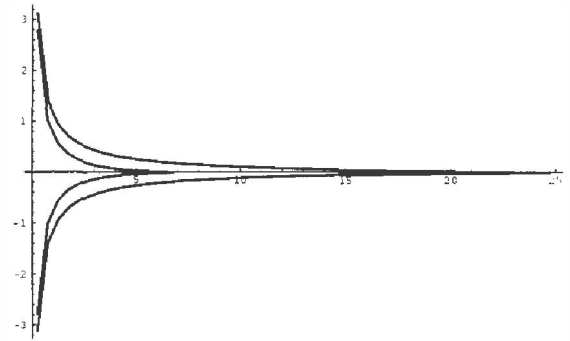

Figure 4: Decision regions for $c_i = 0.02$.

The plans have a satisficing character in which they are admissible in the light of prior beliefs, yet they are open to challenge: during operation, the robustness of any decision may be compared to other admissible decisions and better courses of action can be learned or experimented. There is no need to obtain a single "best" plan and then conduct sensitivity analysis on it: the overall strategy already encodes all robust and non-robust situations the agent may face. Since the credal set and the decision regions are available to the agent, robustness questions can be dealt with in a straightforward manner. From this simple example we notice how the questions of model precision, computational effort and robustness can be all tied together in the New Quasi-Bayesian framework.

## 5 CONCLUSION

We have proposed a new approach to decision-making with Quasi-Bayesian models. Quasi-Bayesian theory maintains that a convex set of probability distributions captures the beliefs of an agent. The theory does not specify how two decisions are to be compared when they are both admissible. This has led to a great deal of anxiety among researchers, who have proposed additional constraints to allow any comparison to be made. We depart in a somewhat radical way from this tradition: instead we propose that any admissible decision can be chosen, but that robustness must be monitored by the agent. A non-robust decision must be refined if possible. If there is no time for refinement, a default admissible action is used.

We have demonstrated this approach to decision-making in a problem of planning to observe, where prior beliefs are captured by a set of Gaussian distributions. We demonstrate how to monitor robustness and how to choose E-admissible actions. Our proposal can be extended to the class of planning to observe problems with multivariate data or with general distributions and conjugate priors (for example, beta priors with binomial observations [Lindley and Barnett, 1965]). Further research is needed to extend these ideas to other, more general decision problems.

The plan generation algorithm here developed is, to the best of our knowledge, the first example of a situation where Quasi-Bayesian theory helps to reduce the complexity of generating a decision plan. This is due to our focus on the robustness, rather than the optimality, of a solution. We expect this approach to shed light on the relationship between rationality requirements and computational effort. Note that we *do not* suggest that models should be imprecise to facilitate search. We *do* suggest that the use of a model should be compatible with its precision. The Bayesian strategy sometimes seems excessive in that it forces a precise model into a problem and then demands optimality or meta-analysis with respect to that model. A Quasi-Bayesian approach that focuses on robustness and computational effort can offer a new perspective for decision making.

The theory developed above admits a different, possibly fruitful, interpretation. Suppose an agent has a Quasi-Bayesian model and the agent is not interested in the robustness of actions; instead, the agent wishes to generate admissible actions as fast as possible. This interpretation of Quasi-Bayesian decision making (as advocated by [Good, 1983]) is that the agent has exhausted preferences and can pick admissible actions arbitrarily. We demonstrated that, for the planning to observe problem, the agent can generate E-admissible plans faster than a Quasi-Bayesian agent could generate a "best" plan.

## A  THE QUASI-BAYESIAN RISK

We wish to minimize the Bayes risk for a Gaussian prior with mean $\mu$, variance $1/\tau$ by using a plan $\delta$. The Bayes risk is

$$\rho(\mu, \tau, \delta) = E[L(\theta, \delta(X_1 \ldots X_n)) + nc].$$

Note that the number of observations $n$ is also a random variable to be averaged in the expectation. Call $\rho(\mu, \tau)$ the value of the Bayes risk for the Bayesian best plan.

Dynamic programming applied to this minimization problem leads to a value iteration algorithm [DeGroot, 1970]. Very briefly, the algorithm assumes that two initial guesses of $\rho$ are given: $\tilde{\rho}'_0(\mu, \tau)$ and $\tilde{\rho}''_0(\mu, \tau)$, such that $\tilde{\rho}'_0(\mu, \tau) \leq \rho(\mu, \tau) \leq \tilde{\rho}''_0(\mu, \tau)$.

Two iterations compose the algorithm, one for $\tilde{\rho}'_0(\mu, \tau)$ and another for $\tilde{\rho}''_0(\mu, \tau)$. Each iteration is (primes are dropped since the next expression can be used both for $\tilde{\rho}'_0$ and $\tilde{\rho}''_0$):

$$\tilde{\rho}_{i+1}(\mu, \tau) = \min\left(\frac{\Omega(\sqrt{\tau}|\mu|)}{\sqrt{\tau}}, E\left[\tilde{\rho}_i\left(\frac{\tau\mu + ry}{\tau + r}, \tau + r\right)\right] + c\right). \quad (3)$$

The following fact is guaranteed for any $i$ [DeGroot, 1970] (intuitively, the algorithm "sandwiches" $\rho(\mu, \tau)$):

$$\tilde{\rho}'_i(\mu, \tau) \leq \tilde{\rho}'_{i+1}(\mu, \tau) \leq \rho(\mu, \tau) \leq \tilde{\rho}''_{i+1}(\mu, \tau) \leq \tilde{\rho}''_i(\mu, \tau).$$

The result is:

- if $\tilde{\rho}'_i(\mu, \tau) = \rho_0(\mu, \tau)$ and $\mu \geq 0$, then $(\tau, \mu)$ is in a **Stop1** region;
- if $\tilde{\rho}'_i(\mu, \tau) = \rho_0(\mu, \tau)$ and $\mu \leq 0$, then $(\tau, \mu)$ is in a **Stop0** region;
- if $\tilde{\rho}''_i(\mu, \tau) \neq \rho_0(\mu, \tau)$, then $(\tau, \mu)$ is in the **Continue** region.

This produces the decision regions, with the **Continue** region between the **Stop0** and **Stop1** regions. Intuitively, the algorithm "sandwiches" the **Indeterminate** region.

In order to start value iteration, the following choices are adequate: $\tilde{\rho}'_0(\mu, \tau) = 0$ (always smaller than $\rho(\mu, \tau)$), and $\tilde{\rho}''_0(\mu, \tau) = \rho_0(\mu, \tau)$, where $\rho_0(\mu, \tau) = \frac{\Omega(\sqrt{\tau}|\mu|)}{\sqrt{\tau}}$. The function $\rho_0(\mu, \tau)$ is always larger than $\rho(\mu, \tau)$ [DeGroot, 1970].

## B  A PACKAGE FOR QUASI-BAYESIAN PLAN GENERATION

The results discussed in this paper were implemented in a Mathematica$^{TM}$ package which is publicly available through anonymous ftp. Connect to **ftp.cs.cmu.edu** as anonymous, go to the directory **/afs/cs/project/lri-3/ftp/outgoing/** and get the file *quasi-bayes.tar*. Use the **tar** program and read the *README* file for the necessary guidance.

## Acknowledgements

This research is supported in part by NASA under Grant NAGW-1175. Fabio Cozman is supported under a scholarship from CNPq, Brazil.

We thank Prof. T. Seidenfeld and L. Chrisman for reading an earlier draft and suggesting substantial improvements to this work.